# Design of Conversational Humanoid Robots based on Hardware Independent Gesture-generation

Katsushi Ikeuchi, David Baumert, Shunsuke Kudoh, and Masaru Takizawa

***Abstract*— With an increasing need for elderly and disability care, there is an increasing opportunity for intelligent and mobile devices such as robots to provide care and support solutions. In order to naturally assist and interact with humans, a robot must possess effective conversational capabilities. Gestures accompanying spoken sentences are an important factor in human-to-human conversational communication. Humanoid robots must also use gestures if they are to be capable of the rich interactions implied and afforded by their humanlike appearance. However, present systems for gesture generation do not dynamically provide realistic physical gestures that are naturally understood by humans. A method for humanoid robots to generate gestures along with spoken sentences is proposed herein. We emphasize that our gesture-generating architecture can be applied to any type of humanoid robot through the use of labanotation, which is an existing system for notating human dance movements. Labanotation's gestural symbols can be computationally transformed to be compatible across a range of robots with differing physical characteristics. This paper describes a solution as an integrated system for conversational robots whose speech and gestures can supplement each other in human-robot interactions.**

## I. Introduction

THE service sector is a growing application field for robotics. Robots that can communicate verbally are receiving increased attention as a research subject and as potential solutions for the service sector. The cost of humanoid robots is expected to decrease until eventually users will be able to afford them for assistance in their daily needs. Therefore, conversation-capable robots that can also provide other physical services, such as elder care or daily personal assistance, should be developed.

Conversational capabilities are a key requirement for service robots to coexist harmoniously with humans. When robots are built with humanoid forms, it follows as a natural expectation they will also exhibit human-like behavior. Thus, several studies have attempted to develop robot designs that support conversational interactions with humans in real-life situations. Nishio et al. developed a teleoperated system for android robots [1]. Shiomi et al. developed a semiautonomous conversational system that reverts to a human operator only when the robot encounters a situation it cannot handle by itself [2]. This past work shows how human operators can enable robots to respond appropriately to unexpected situations. However, for practical uses that demonstrate the real value of robotics, conversational systems need to be fully autonomous.

Conversation agent software applications, often referred to as chatbots, are computer programs written to conduct autonomous conversations with humans. These conversational agents have been refined over time and their performance has improved with new machine-learning programming techniques, the increased availability of training data, and increasing computer hardware performance. Bessho et al. developed a system that autonomously creates responses using real-time crowdsourcing [3]. Higashinaka et al. proposed a system for improving an utterance-generation system using Twitter's large-scale dataset [4]. These systems have been implemented in smartphones to provide assistance functions [5]. However, few studies have discussed the effectiveness of embodied robots with integrated chatbot systems.

Generating appropriate gestures along with verbal expressions is an important issue when combining chatbot systems with embodied agents or robots [6]. If the robot's conversational content is limited, an appropriate gesture can be designed for each of the robot's utterances. However, with an autonomous conversational system, the robot must generate gestures in real-time for an unlimited set of utterances. Regarding gesture generation for robots, Ishi et al. developed formant-based lip motion generation for teleoperated robots [7]. Sakai et al. generated robot's head motion from linguistic and prosodic information extracted from speech signals [8]. However, these studies did not consider semantics and focused exclusively on the facial and head movements of androids designed to resemble realistic human beings.

Automatic gesture generation for embodied conversational agents has been studied and developed for virtual computer-generated characters [9]. Gestures are generated using two

This paper was submitted for review on April 2nd, 2019. Part of this work was supported by Microsoft Research Asia CORE Project.

Katsushi Ikeuchi and David Baumert are with the Microsoft AI+Research Division, Redmond, WA 98052 USA (email: (katsuike, dbaum)@microsoft.com).

Shunsuke Kudoh and Masaru Takizawa are with Graduate School of Information Systems, the University of Electro-Communications, Tokyo, Japan (email: (s-kudo,takizawa)@is.uec.ac.jp

main methods: a prosody-based generation system [10] and a semantic-based generation system [11]. Marsella et al. reported that by considering both prosody and semantics, more appropriate gestures can be generated for virtual characters as compared to a prosody-only based system [12]. Kippe et al. proposed a system to generate gestures based on probabilistic reproduction of human behaviors [13]. These systems perform well when designed for a virtual character but cannot be applied to physical robots due to the differences between virtual models and the hardware performance and physics-based constraints presented by physical robots. Also, it is difficult to directly transfer gestures that work on one robot to a physically different robot. Even if they share the basic humanoid form, hardware differences such as limb length and actuator placement change the physical characteristics of the same gesture across different robots. In this paper, we propose a system that overcomes this machine dependency by separating the hardware-independent and hardware-dependent components used to generate gestures.

This paper aims to present an integrated system architecture with two components that provides a solution for a humanoid robot to generate gestures aligned with its speech. The first component is a gesture library comprised of gesture-concept pairs (See Fig. 1). It is generally accepted that in human-to-to human verbal communication, gestures accompanying speech help to elucidate underlying concepts [14]. The gesture library employs the principle that a group of words can be paired with a group of gestures that represent the same concept [15]. The gesture library can be stored locally in the robot or in a networked cloud service. The second component is a gesture engine that receives an utterance to rendered by the robot from a conversational agent component and then retrieves a corresponding gesture-concept pair from the gesture library. This component, as well as the conversational agent, can be run locally on the robot or as a networked cloud service. The pair is selected by analyzing the utterance to determine the underlying concept and finding a match in the gesture library. The gesture paired with the concept is then delivered to the robot's actuator-control system as a generalized representation described in labanotation. Finally, the robot executes software adapted to transform the labanotation into actuator commands appropriate for that robot's specific hardware configuration.

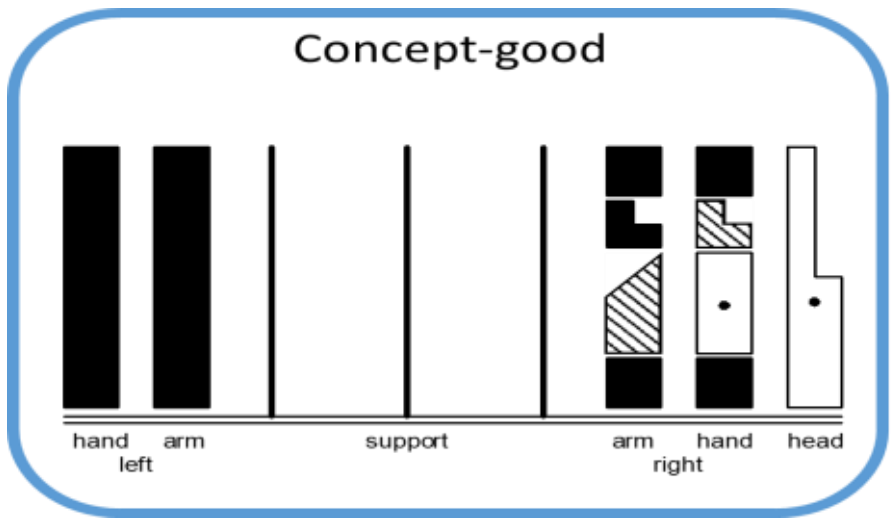

Fig. 1. An example concept-gesture pair. One pair comprises one gesture and one corresponding concept. Each gesture is described using a symbolic representation, Labanotation.

The remainder of this paper is organized as follows: Section II presents the gesture library, Section III presents the gesture engine, and Section IV describes an example implementation. Section V discusses progress, limitations, and opportunities for future study.

## II. Gesture Library

According to McNeill, a gesture and a word can share the same concept when rendered by a speaker [14]. This idea inspires us to create a database of gesture-concept pairs, which we refer to as a gesture library. Our conjecture is the number of gesture variations is smaller than the number of concept variations; thus, we index the library based on gesture variations.

The first step for creating the library is to collect data describing the various gestures used in human speech. As a source of sentences and utterances, we employed an English textbook containing 230 representative conversations. We asked human subjects to conduct these conversations with their own gestures. A special instruction is given so that human subjects should avoid deictic, question and beat gestures by using some alternative gestures so that we can increase the variety of gestures recorded. These gestures are performed in front of an optical sensor system that captures movements based on skeletal structure. The motions of the subject are recorded as sequences of two-dimensional color and three-dimensional depth images and then processed into sequences of stick figures.

The stick figure sequences are converted into labanotation, an existing system that lends itself to a machine-independent representation of gestures. The labanotation system was first proposed by Rudolf von Laban as a method to describe human gestures in symbolic representations for the purpose of documenting choreography in the dance community [16]. Fig. 2(a) shows the example of a Labanotation score. The columns of the diagram correspond to each body segment as shown in Fig. 2(b). Time flows from bottom to top. Each symbol in the column indicates one of 28 digitized directions as indicated in Fig 2(c), and the length of a symbol depicts the duration of one movement of the body segment.

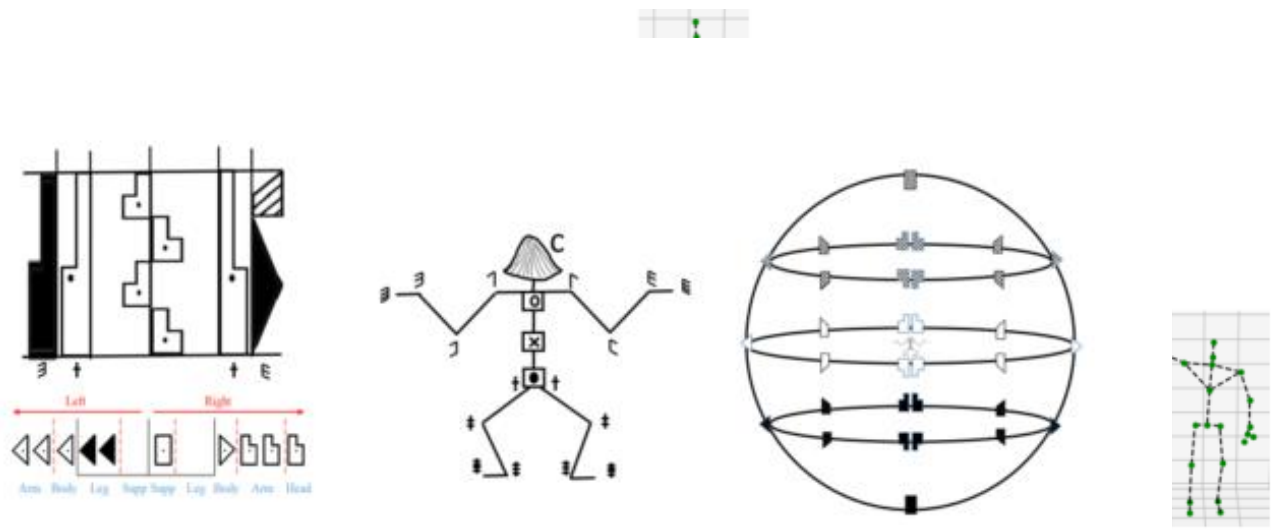

Fig. 2. Example of Labanotation; (a) A Labanotation score. The columns correspond to each body segment, which is indicated using an arrow-like symbol. The shape and its color of each symbol represents a direction at a key pose. The length of a symbol depicts the duration of one movement, an interval of two key poses. (b) Body segments with arrow-like symbols. (c) 28 spatial directions depicted on the Gaussian sphere. The azimuth direction is depicted with a symbol shape and the zenith direction is depicted with a symbol color.

The relationship between a labanotation score and the performance of a dance is parallel to the relationship between a music score and the performance of a music piece. In an idealized context, different performances of the same music piece would be recorded as the same musical score. Similarly, different performances of the same dance would be recorded as the same labanotation score. As a result, even with some differences between performers and performances, observers of dance or listeners of music perceive the rendered pieces to be the same.

We employ the 3D Robot Laban Suite [17] to obtain labanotation scores from sequences of stick figures as shown in Fig. 3. That system converts stick figure movements into velocity values for each body segment and places them on a spherical coordinate system to obtain the local minima points of the velocities, corresponding to brief stops, for each body segment. The brief stops along the time sequence are referred to as key frames. At each key frame, corresponding to the boundaries of symbols in the labanotation score, the directions of each body segments are collated into one of twenty-eight distinct directions. This digitized result represents a labanotation score.

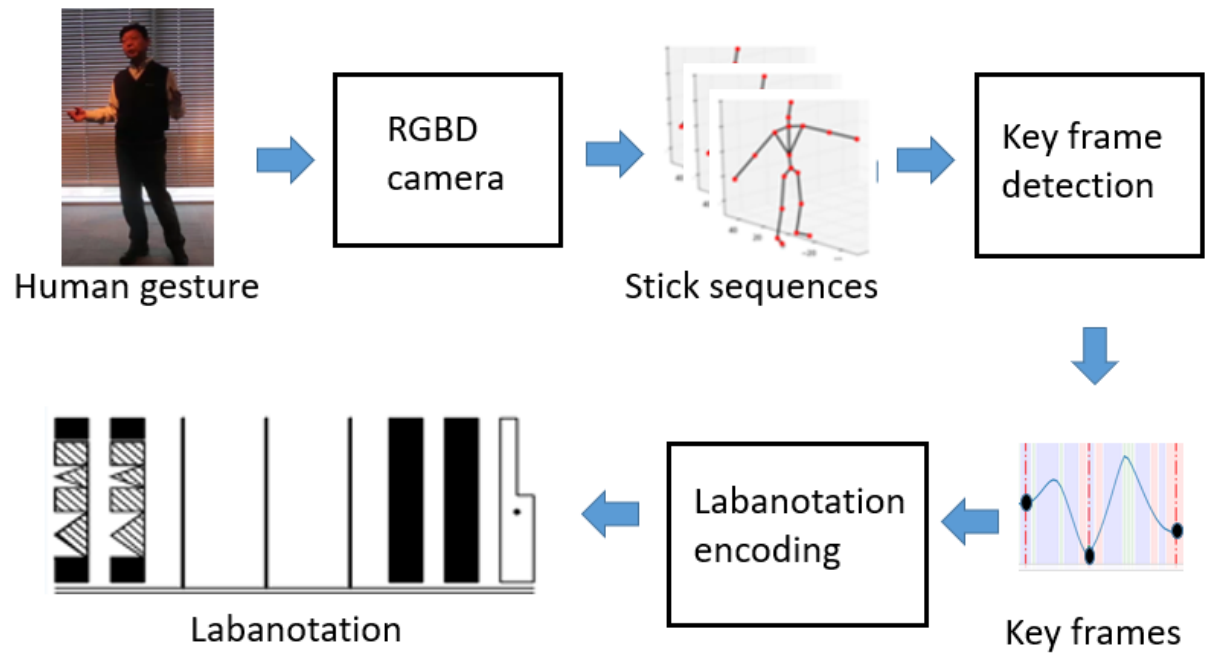

Fig. 3. 3D Robot Laban Suite – gesture to Labanotation conversion system [17]. A human gesture is recorded by a RGBD camera, and then converted into a stick sequence. Brief stops along the time sequence, referred to as key frames are detected as the local minima points of the velocities of body segments. At each key frame, the directions of each body segment are collated into one of the twenty-eight directions. This process provides the Labanotation score of the gesture.

Fig.4 shows one example of a Labanotation score obtained by the 3D robot Laban Suite and corresponding the stick figures to each Labanotation symbols.

A Labanotation score of one gesture, L, is represented as a matrix, where columns and rows correspond respectively to body parts and timings. Each element of the matrix represents the body segment direction digitized into twenty-eight directions. Since we limited our scope to the upper body in our conversational robot, there are only five body segments to consider. Therefore, given T sampling along the time axis, each gesture is represented as a T X 5 matrix:

$$L = \begin{bmatrix} l_{T,l-lower} & l_{T,l-upper} & l_{T,r-lower} & l_{T,r-upper} & l_{T,head} \\ \vdots & \vdots & \vdots & \vdots & \vdots \\ l_{2,l-lower} & l_{2,l-upper} & l_{2,r-lower} & l_{2,r-upper} & l_{2,head} \\ l_{1,l-lower} & l_{1,l-upper} & l_{1,r-lower} & l_{1,r-upper} & l_{1,head} \end{bmatrix}. \quad (1)$$

The similarity between two gestures is measured as the distance between two Labanotation scores, L and M, calculated as the summation of geodesic distances between each element.

$$D = \sum_{i=1}^{T} \sum_{j=1}^{5} d\,(\,l_{ij}\,, m_{ij}\,) \quad . \quad (2)$$

Here, $d\,(\,l_{ij}\,, m_{ij}\,)$ is the geodesic distance along the Gaussian sphere of two directions $l_{ij}$ and $m_{ij}$. Using this measure, 230 Labanotation scores were grouped into 32 clusters. Each cluster consists of words that share a similar Labanotation score.

At each cluster, we manually identify and extract a common concept. This common concept is utilized as the name of the cluster. During experimental demonstrations of the system, a trend was revealed where the repetition of exactly the same gesture for a given repeated concept was perceived as unnatural by the audience. To compensate, we add small variations to gestures from heavily-utilized clusters.

Fig. 5 shows three pairs out of the thirty-two clusters as an illustrative example of concept-gesture pairs, while the Appendix contains all of the gesture clusters. The first column denotes the cluster name. The second column represents words representing the gesture cluster. The third column represents the labanotation score of the gesture. The fourth column contains an image sequence of a robot performance of the gesture.

## III. Gesture Engine

### A. *Outer structure*

The outer structure of the gesture engine is a rule-based system. Some responses from a conversation engine are very short, such as the sentence, “hi”. We limit one gesture corresponding to one incoming sentence so that we can guarantee any robot to be able to perform such gesture in an open loop manner without any hardware stack due to the operation collision. Namely, the gesture engine is designed to select one and only one gesture from the library based on the rule.

The first operation is to divide the sentence to be rendered by the robot into words:

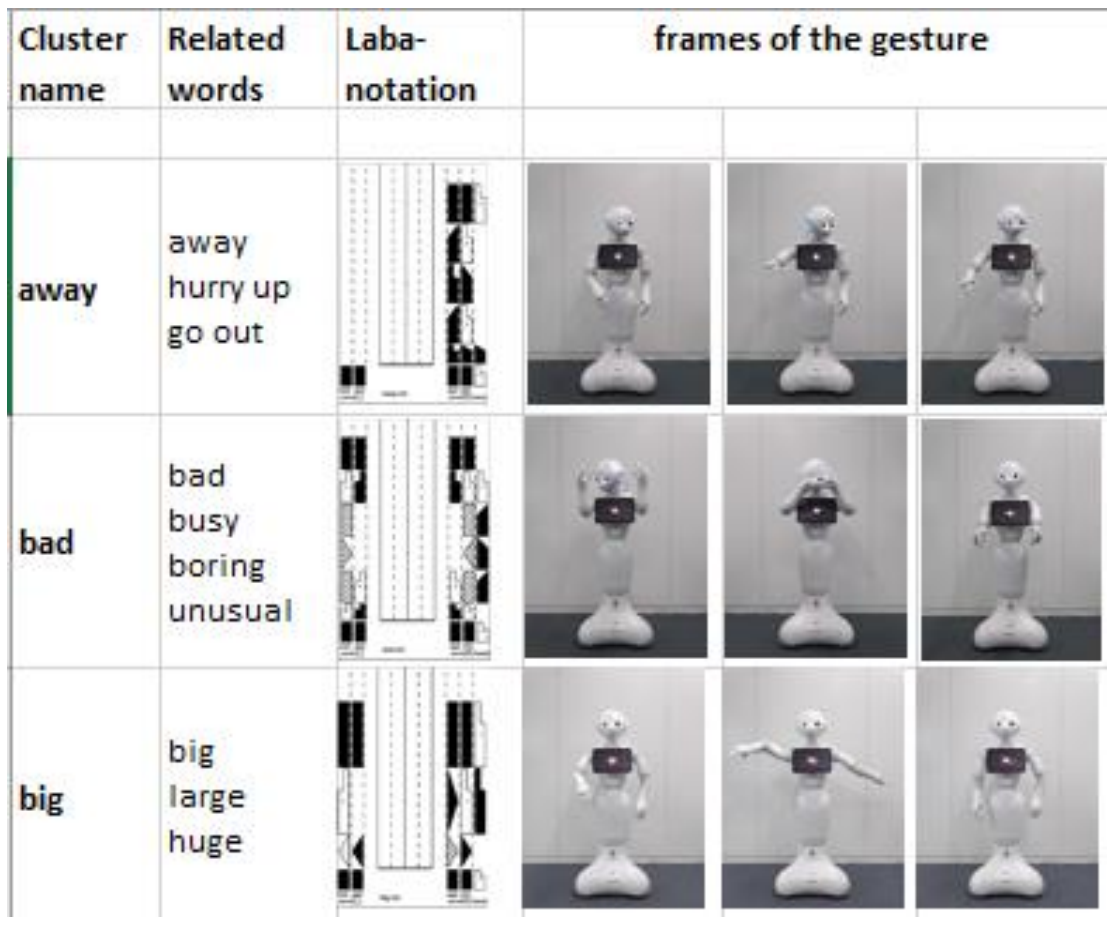

Fig. 5. The first three of the thirty-one concept-gesture pair clusters.

$$S = (S_1, S_2, \cdots, S_n)\ . \tag{3}$$

Here, each component corresponds to one word.

Next, the engine determines if any word in the sentence is deictic by performing a pattern match between S and the list of words: *this, that, here* and *there*. If a deictic word is found in $S_1$, a random number between 0 and 10 is generated. If the number is larger than five, the matching gesture-concept pair is selected from the gesture library and the corresponding labanotation score is retrieved. This random process is introduced to avoid the repletion of deictic gestures. For this sentence, the matching process is complete, and the labanotation score along with the sentence text are sent to the edge computer for performance rendering. If the random number is smaller than the threshold, the rule-based system ends the deictic part and proceeds to the next step for the gesture search.

The second step of the rule-based system is to determine if it contains a questioning word. A pattern match between $S_1$ and the list of words: *who, what, when, where* and *how*. If a questioning word is found in $S_1$, the random test is executed. Other questioning words such as *Do* or *Does* are intentionally ignored. Question sentences containing these words are often accompanied by relatively small and quick question gestures at the end of the sentence. Since our design is open-loop, we cannot assume any synchronization between voice sounds and motor control signals.

If the test is passed, the matching gesture-concept pair is selected from the gesture library and the corresponding labanotation score is retrieved. For this sentence, the matching process is complete, and the labanotation score along with the sentence text is sent to the edge computer for open-loop performance rendering. Otherwise, the third step is executed.

The third step determines similarity between the incoming sentence and the all the concepts in the gesture library by using the word-vector projection, which will be described in the next section in detail [18]. A threshold is set to determine a valid match. If the maximum similarity is higher than the threshold, candidate gesture-concept pairs within 95% of maximum similarity are selected. And, then, among the obtained pairs, one pair is randomly selected to avoid repetition.

If the maximum similarity is lower than the threshold, the gesture-concept pair for “beat” is selected from the library as a default gesture.

### *B Pair Similarity Based on Word-Vector Space*

In the case where the gesture engine selects an appropriate concept-gesture pair from the gesture library using word-vector projection, we use this process:

First, recall that a group of gesture-concept pairs in the library will contain representative words:

$$G = (g_1,\ g_2, \cdots, g_m)\ . \tag{4}$$

The spoken sentence is represented as a list, $S = (S_1, S_2, \cdots, S_n)$, where $S_i$ depicts one word and *n* is the word length of the sentence. We define the similarity between a spoken sentence, *S* and one gesture group *G* as follows: First, the words in the sentence are converted into vectors in a low-dimensional vector space [18]. Then, the similarity between two words is defined based on their cosine distance in that same low-dimensional space. Next, the similarity of each word, $S_i$ to the gesture group is $(g_1,\ g_2, \cdots, g_m)$ considered to find the maximum similarity between the input word and the words found in the group of gesture-concept pairs:

$$f(s_i) = \max_j similiarity\ (\ s_i\ , g_j\ )\ . \tag{5}$$

Finally, the similarity between the spoken sentence and one gesture group, *T*, is defined as the summation of the above similarity as follows:

$$T = \sum_{i=1}^{n} g(f(s_i)). \tag{6}$$

Here, g is a function used to determine the selection of dissimilar groups and to encourage random selection among similar gestures, which is defined as follows:

$$g(e) = \begin{cases} 0 & e < \tau_0 \\ \frac{e-\tau_0}{\tau_1-\tau_0} & \tau_0 \le e \le \tau_1 \\ 1 & \tau_1 < e \end{cases}\ , \tag{7}$$

where $\tau_0$ and $\tau_1$ are empirically defined thresholds.

## IV. Implementation

This section provides an overview of an example system using a cloud/edge architecture (Fig. 6). The cloud software is designed to be a machine-independent implementation and the edge software handles machine-dependent transformations based on the robot's specific physical characteristics. Speech input representing a user's question is directly sent to the cloud and converted into text format. The user's question in text format is sent to conversational agent software running in the cloud. The conversational agent generates an answer to the user's question. The answer in text form is given to the gesture engine also running in the cloud. The gesture engine uses its logic to select a gesture-concept pair from the gesture library which also resides in the cloud. Then, the gesture engine sends back the data consisting of the text.

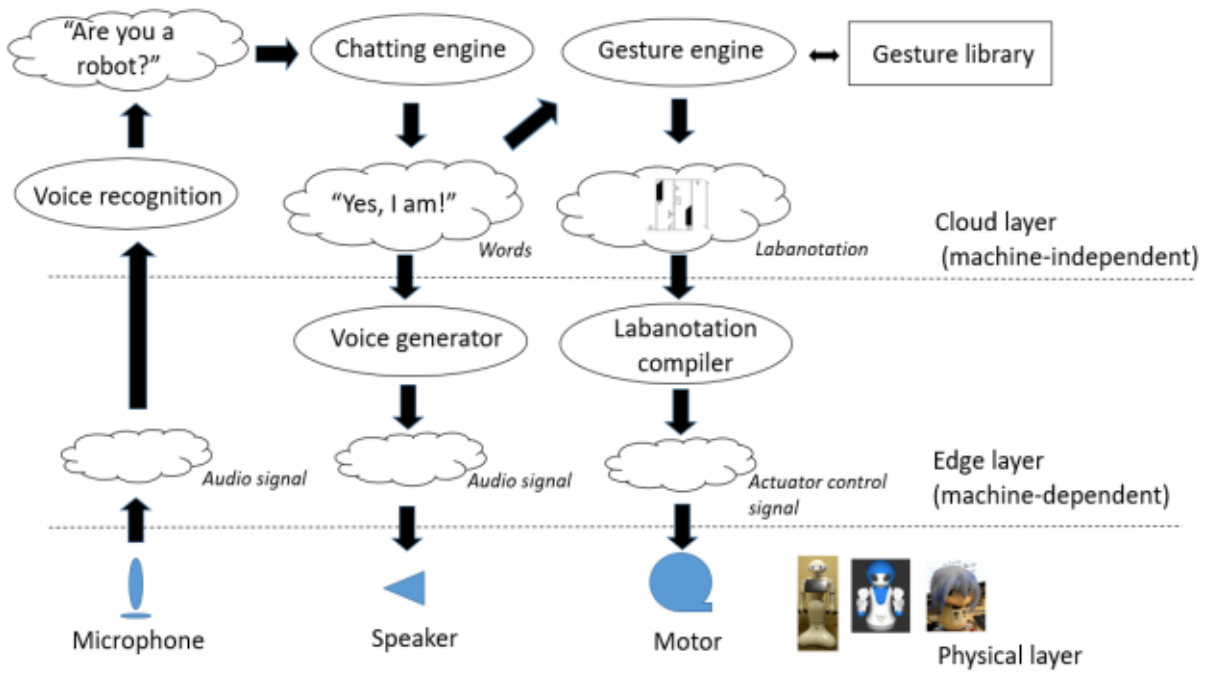

Fig. 6. Overview of gesture-generations system using a cloud/edge architecture.

At the edge side, the received text is converted into an audible speech signal and the labanotation score is converted to robot actuator control signals. The speech and control signals are sent to the robot's speaker and joint actuators. In our prototype implementation, we employed Softbank's Pepper device and our in-house robot, MS-RABOT as the robot platforms. In subsequent implementations of the system, the merit of using the labanotation score is obvious when changing robot hardware. It was only necessary to update the compiler that converts labanotation to motor control signals running on the robot. This edge implementation is straightforward when the degree of freedom (DOF) of the target robot is the same as a human. In the case where the DOF does not match, a set of conversion rules is required [17]. The cloud components require no change. Since the data contains both texts to be spoken and Labanotation to be executed, if a robot has a capability to synchronize them, the Labanotation compiler will synchronize their execution. If not, audio generation and motor execution are performed in an open loop manner.

## V. Conclusion

This paper proposes a machine-independent gesture-generating architecture for conversational robots and evaluates the validity of our method via a prototype implementation. Interactions with the prototype system showed that gestures corresponding to un-planned text sentences can be consistently found and rendered by different humanoid robots. This shows that our proposed gesture engine can successfully consider the semantic meaning of a sentence in making the choice of an appropriate corresponding gesture from a gesture library. Furthermore, it shows that labanotation can be the basis of gestural behavior provided to robots as a general and intelligent service, relieving us from the requirement to create a custom gesture design for each new humanoid robot. This implies that gestures can be considered to play a supplemental role in improving the natural quality of verbal interactions between humans and robots.

In future work, we would like to develop autonomous generation of gesture libraries, where humanoid robots autonomously learn the relationship between speech and gestures through conversations to humans and enrich gesture libraries.

## APPENDIX

Current gesture library contains thirty-two concept-gesture pairs in the general gesture category, four deictic, one question and one beat pairs. Fig. A1 contains concept-gestures of deictic, question and beat gestures, while Fig.A2 contains thirty-two general pairs.

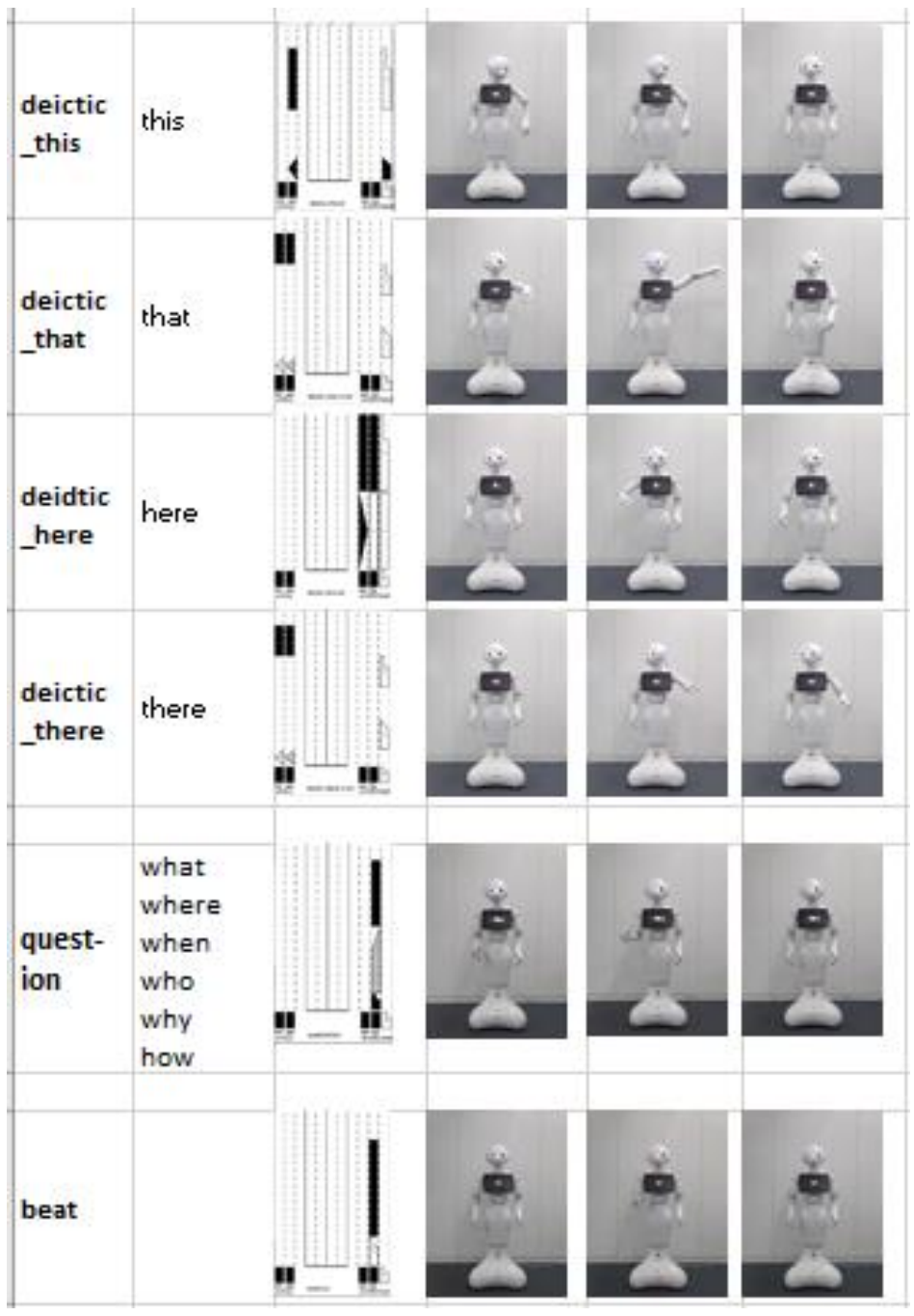

Fig. A1 deictic, question and beat pairs

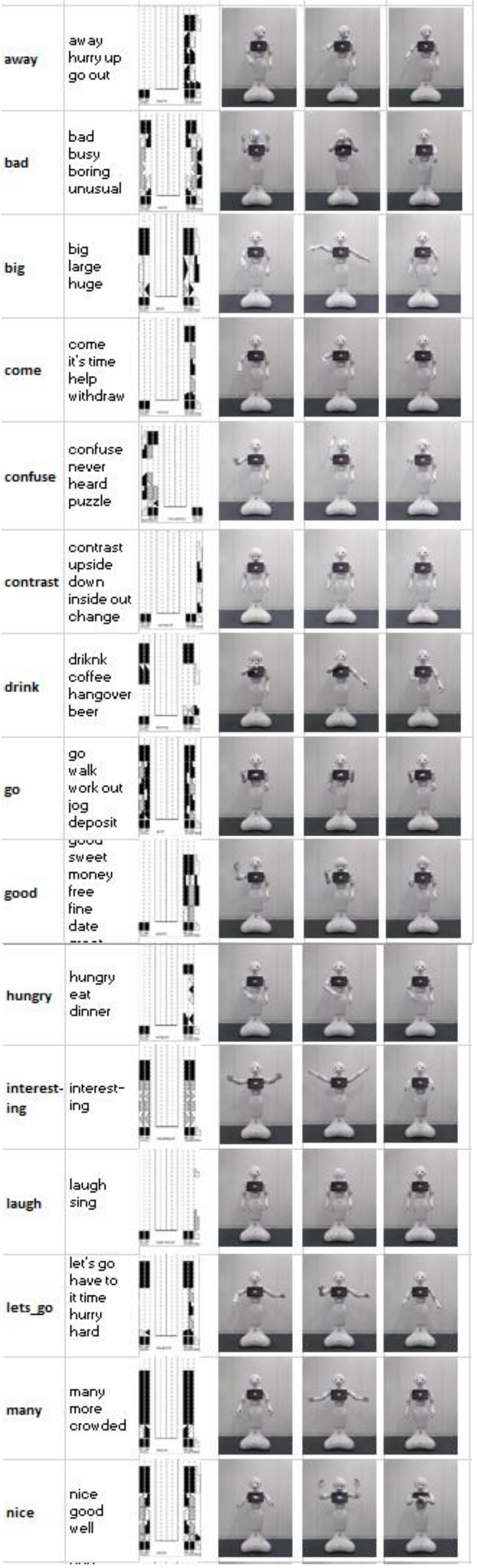

Fig. A2 General pairs (cont.)

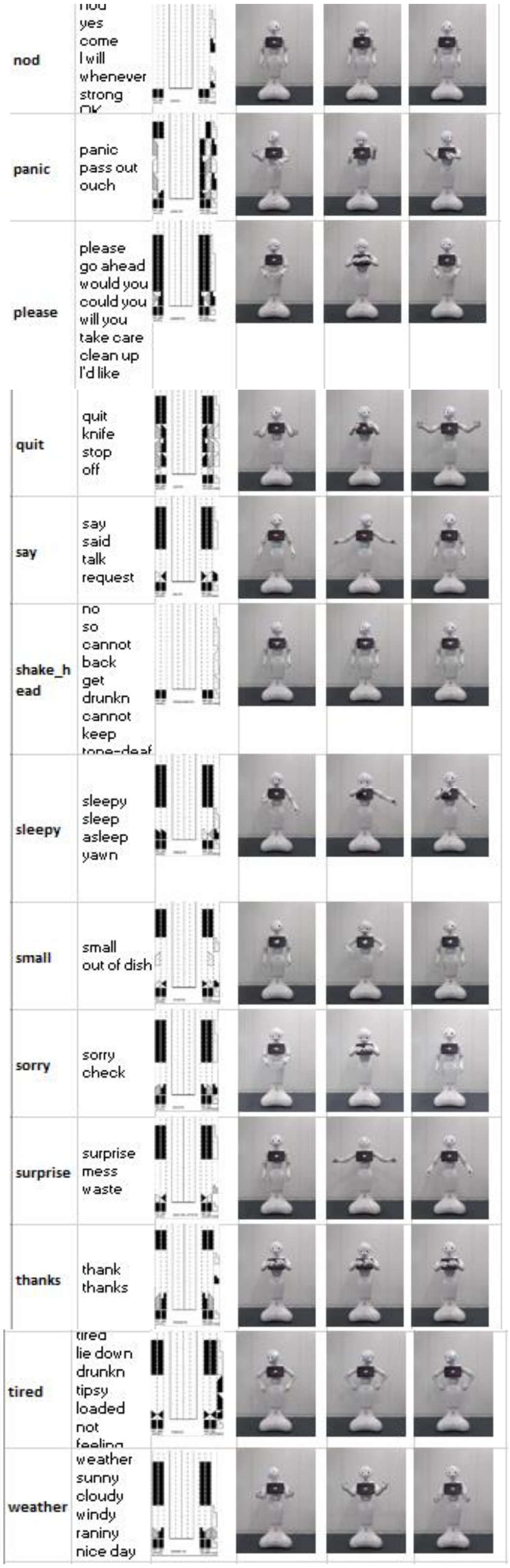

Fig. A2 General pairs